\pdfoutput=1
\documentclass[letterpaper, 10 pt, journal, twoside]{IEEEtran}  

\IEEEoverridecommandlockouts                              

\usepackage{algorithm}
\usepackage{multicol}
\usepackage{multirow}
\usepackage{algpseudocode}
\usepackage{color}
\usepackage{xcolor}
\usepackage{graphicx, epstopdf}
\usepackage{bm}
\usepackage{amsmath}
\usepackage{amssymb}
\usepackage{tabularx}
\usepackage{subcaption}
\usepackage{url}
\usepackage{cite}
\usepackage{prettyref}
\usepackage{booktabs}
\usepackage{siunitx}
\usepackage{tikz}
\newcommand*{\circled}[1]{\lower.7ex\hbox{\tikz\draw (0pt, 0pt)%
    circle (.45em) node {\makebox[1em][c]{\small #1}};}}
\usepackage{amsfonts}

\usepackage{verbatim}

\usepackage[capitalize]{cleveref}
\crefname{section}{Sec.}{Secs.}
\Crefname{section}{Section}{Sections}
\Crefname{table}{Table}{Tables}
\crefname{table}{Tab.}{Tabs.}

\makeatletter
\newcommand{\printfnsymbol}[1]{%
  \textsuperscript{\@fnsymbol{#1}}%
}
\makeatother

\title{
DiffSRL: Learning Dynamical State Representation for Deformable Object Manipulation with Differentiable Simulation
} 

\author{Sirui Chen$^{*1}$, Yunhao Liu$^{*1}$, Shang Wen Yao$^1$, Jialong Li$^1$, Tingxiang Fan$^{1,2}$, Jia Pan$^{\dagger1,2}$%
\thanks{Manuscript received: April, 30, 2022; Accepted July 27, 2022.}
\thanks{This paper was recommended for publication by Editor Jens Kober upon evaluation of the Associate Editor and Reviewers' comments.
This work was supported by HKSAR Research Grants Council (RGC) General Research Fund (GRF) HKU 11202119, 11207818, and the Innovation and Technology Commission of the HKSAR Government under the InnoHK initiative. S. Chen, Y. Liu, S. Yao are also supported by Undergraduate Research Fellowship Program (URFP) of HKU.}
\thanks{$^*$ denotes equal contribution. $^\dagger$ denotes the corresponding author.}
\thanks{$^1$ S. Chen, Y. Liu, J. Li, S. Yao, T. Fan, J. Pan are with the University of Hong Kong. 
    {\tt\footnotesize jpan@cs.hku.hk}}%
\thanks{$^2$ T. Fan and J. Pan are also with the Centre for Garment Production Limited, Hong Kong.}
\thanks{Digital Object Identifier (DOI): see top of this page.}
}

\begin{document}
\maketitle

\begin{abstract}


Dynamic state representation learning is essential for robot learning. Good latent space that can accurately describe dynamic transition and constraints can significantly accelerate reinforcement learning training as well as reduce motion planning complexity. However, deformable object have very complicated dynamics and is hard to be represented directly by a neural network without any prior physics information. We propose DiffSRL, an end-to-end dynamic state representation learning pipeline that uses differentiable physics engine to teach neural network how to represent high dimensional pointcloud data collected from deformable objects. Our specially designed loss function can guide neural network aware physics constraints and feasibility. We benchmark the performance of our methods as well as other state representation algorithms with multiple downstream tasks on PlasticineLab. Our model demonstrates superior performance most of the time on all tasks. We also demonstrate our model's performance in real hardware setting with two manipulation tasks on a UR-5 robot arm. The source code is available at \url{https://github.com/Ericcsr/DiffSRL/} and our attached video.

\end{abstract}


\section{Introduction}
\label{sec:intro}

Deep neural networks have become a powerful tool in representing high dimensional data using fewer dimensions. Nowadays, its rapid development and application has significantly enhanced the machines to process complex data such as images, sentences and graphs. One application that has recently aroused great research interest is representing real-world objects in dynamic environments from high dimensional point clouds or images for control. Low dimension latent space can benefit learning algorithms by reducing the tunable parameters, accelerating the training, and achieving better control results. The efficiency of this paradigm on rigid body has been demonstrated in recent research~\cite{E2C, AutoEncoder}.

\begin{figure}[htbp!]
\centering
  \includegraphics[width=0.8\linewidth]{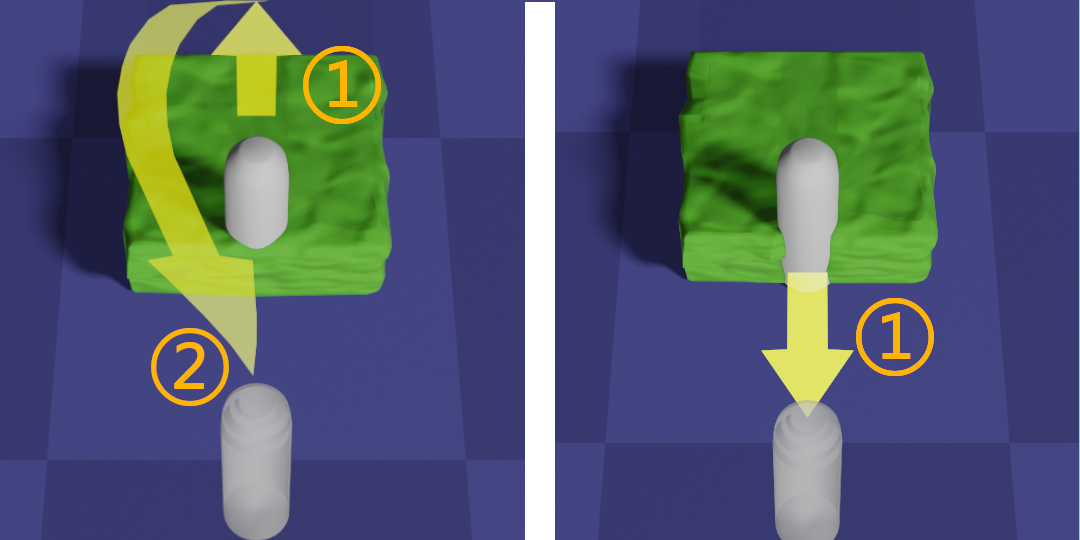}
  \caption{Similar states but expecting distinct action trajectories.}
  \label{fig:similar}
\vspace{-20pt}
\end{figure}

However, representing commonly seen deformable objects in low dimensions remains a challenge~\cite{CFM}. It may be accounted for by:
a) State and dynamic complexity. \textcolor{black}{Deformable objects' states are usually represented by particle systems such as meshes or point clouds}. Their dynamics is usually represented by discretizing complex differential equations on particles. Compared with rigid body, they have much larger state space as well as much more complicate dynamics.
b) Correlation complexity. Different particles of the one deformable object can have very different correlations under different deformation~\cite{CFM}. For instance, all points on the surface of a piece of cloth have the same movement when the cloth is translating as a whole. However, when deforming, the cloth is divided into many parts where surface points move independently.
c) Topological complexity. Because of different topology under deformation, visually similar states may not have same topology, making it confusing for state representation learning methods to distinguish different states basing on static visual metrics. \cref{fig:similar} shows one example where a rod is to be removed from a piece of plasticine and placed at a target position with minimum energy consumption. The two initial states are similar but have subtle differences: the rod is surrounded by plasticine on the left, while there is a notch on the right. Static representation encoders may overlook their difference because the notch occupies a tiny volume of space. Thus, controller may generate similar actions based on latent state from a static state encoder, but we prefer an algorithm that first drags the rod out (action \circled{1}) and then translates it (action \circled{2}) in the left case, and directly translates the rod through the notch in the right one. Similar challenges are ubiquitous in soft object manipulation, calling for new state representation learning algorithms, capable of both examining the current state and understanding the dynamics to reflect different futures bifurcated from similar present, which we defined as the "dynamic-awareness."

Existing state representation learning methods developed from rigid body systems, such as AutoEncoder \cite{AutoEncoder} and Embed to Control (E2C) \cite{E2C}, lack specific designs for the above challenges related to deformable objects. Newly emerged solutions focusing on deformable objects, such as the Contrastive Forward Model (CFM) \cite{CFM} or G-Doom \cite{G-DOOM}, as is elaborated in \cref{sec:related_work}, are very data demanding since they fail to utilize physics prior in their training pipelines, leading to a performance bottleneck on more complicate deformable objects other than ropes or cloth under very limited deformation \cite{SRLSurvey}. To improve the performance of the state representation learning on deformable objects, we propose a new pipeline, DiffSRL, that utilizes a differentiable simulator to encode dynamic and constraint-related information.

Our main contributions are two-fold.
\begin{itemize}
\item \textcolor{black}{We propose an data efficient end-to-end dynamic state representation learning pipeline utilizing differentiable simulation to capture the multistep dynamics of deformable objects. To the best of our knowledge, this is the first time that state representation learning directly uses the simulation as physics prior as a part of the end-to-end training pipeline. We compared different state representation learning algorithms such as CFM\cite{CFM} inverse model learning\cite{Inverse}, AutoEncoder\cite{AutoEncoder} and E2C\cite{E2C} with our method on different downstream tasks varying from simple trajectory reconstruction to reinforcement learning algorithm for manipulation. Our method achieves the best performance among others most of the time.}
\item We design a novel loss function in our proposed method that guides the encoder to be aware of dynamical constraints during training, which can improve the performance of downstream tasks.
\end{itemize}
We also transfer learned policy from the simulator to the real world. We conducted experiment on a real robot to verify our model against noisy data from a depth camera.


\section{Related Work}
\label{sec:related_work}
\begin{figure*}[t!]
\centering
  \includegraphics[width=0.85\linewidth]{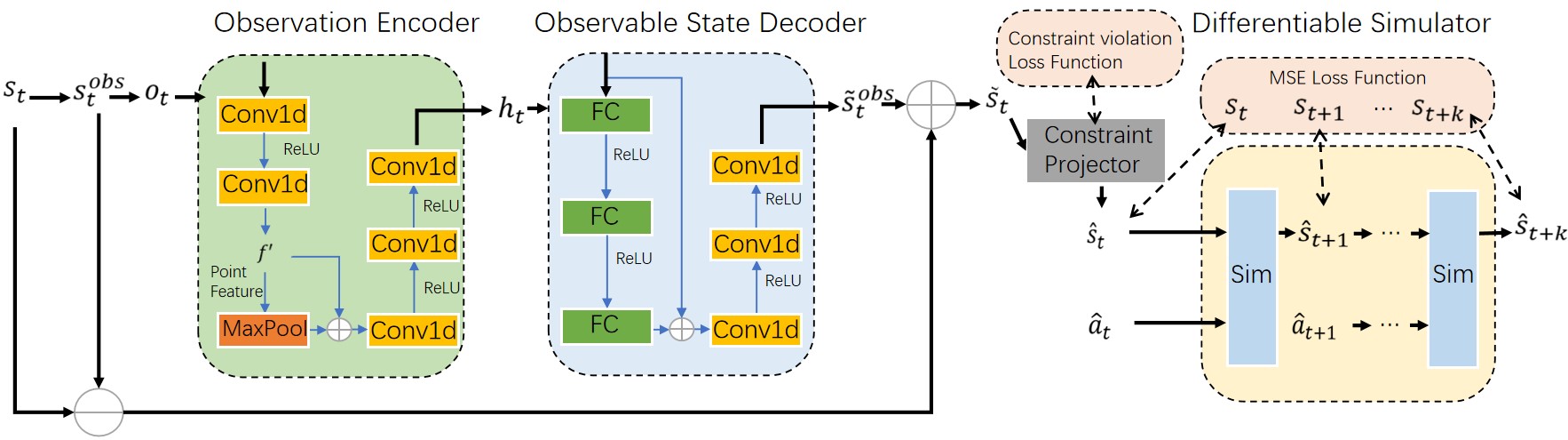}
  \vspace{-8pt}
  \caption{Overview of our DiffSRL model pipeline.
  }
  \label{fig:model}
\vspace{-15pt}
\end{figure*}

One of the most explored methods for state representation learning may be the AutoEncoder~\cite{AutoEncoder}, which trains a neural encoder-decoder network pair to extract features from images to accelerate and enhance the Deep Reinforcement Learning (DRL) by compressing useful information to low-dimension latent state. However it only uses static state information and will fail to distinguish visually similar but dynamically different states~\cite{SRLSurvey} as is discussed above. To address this issue, E2C~\cite{E2C} proposes a pipeline including a learned linear dynamical model together with an AutoEncoder. E2C demonstrates its strength in enhancing DRL’s performances in target-reaching problems with image observations, but its linear model cannot describe deformable objects’ complex dynamics well, especially for long horizon. Replacing the E2C's linear model with a neural network may fail because neural networks with significantly more parameters may be hard to converge with reasonable amount of data and prevent the E2C from achieving desirable outcomes. Alternatively, the inverse-model-based state representation learning~\cite{Duan2017} is established on the heuristics that good state representation should be able to predict actions leading to the transfer between two consecutive states. Thus, the encoder can be trained with an action predictor by minimizing the Mean Square Error (MSE) loss between actual and predicted action. The inverse-model-based method has been further applied on model-free RL tasks ~\cite{LossIsRewards, ICM}, with more variants and applications elucidated in~\cite{SRLSurvey}, but existing designs do not consider the complexity of deformable objects either. CFM~\cite{CFM} explores the state representation learning on deformable objects such as cloth and ropes. It utilizes contrastive learning to jointly train a forward model and an encoder to represent images of deformable objects with distinguishable latent states. The trained forward model and the encoder have been used in MPC-based manipulation control. G-Doom~\cite{G-DOOM} applies a similar paradigm to CFM but with a different architecture in terms of using graphs as the latent space representation and graph-dynamic models as the forward model. However, despite their attempts in handling deformable objects, both CFM and G-Doom don't utilize prior knowledge in physics and require large datasets for relatively simple tasks. 

Neural network encoders are the corner stone for a state representation learning method. Designing a neural network architecture for point cloud is challenging due to its unstructured and order-invariant properties. PointNet~\cite{PointNet} is a pioneering work of applying deep learning on point cloud data, where the permutation invariance is achieved using point-wise symmetric functions, including point-wise Multi-layer Perceptrons (MLPs) and max-pooling layers. Moreover, point cloud encoders that are derived from the PointNet can tolerate different number of input points, allowing it to be evaluated on point clouds with different densities from that of training data. Thus, we use the PointNet as the backbone of our encoder, and utilizes point clouds extracted from the PlasticineLab MPM simulation as the training dataset.

Recent advances of physics-based simulation, such as Material Point Method (MPM) ~\cite{MPM}, allow us to accurately and efficiently simulate deformable objects such as cloth and plasticine. Meanwhile, a new branch of physics simulator, the differentiable simulator~\cite{nimblephysics, DiffHand, DiffTaichi}, make each simulation step differentiable and can be integrated as a layer in deep neural networks, making end-to-end training possible. These simulators provide us with an opportunity to directly introduce physics prior into training pipelines to overcome the difficulty of capturing complex dynamics with a neural net faced by E2C, CFM and G-DOOM. In end-to-end training, gradients from the upstream neural networks are propagated through physics engines and affect network parameters accordingly. For instance, ChainQueen~\cite{ChainQueen} firstly introduces differentiable MPM method to simulate soft robots and build simple controllers. DiffTaichi~\cite{DiffTaichi} and PlasticineLab~\cite{PlasticineLab} extended the differentiable MPM to simulate liquid and plasticine using the Taichi language~\cite{Taichi}. 

\section{Problem Formulation}
\label{sec:problem}
A general dynamic system consists of three components: a state space $\mathcal{S}$, an action space $\mathcal{A}$, and a state transfer function:
\begin{align}
\label{eq:sim_dynamics}
s_{t+1} = f_{\text{sim}}(s_t,a_t),\ s_t,s_{t+1}\in\mathcal{S};a_t\in\mathcal{A}.
\end{align}

The state $s_t$ is obtained from the differentiable simulator which contains physical properties such as coordinates and velocities of particles. To mimic real sensor's observation, the observation $o_t$ contains only the positions of a subset of the particles (we call position of all particles as observable state $s_t^\text{obs}$, so $o_t\in s_t^\text{obs}$). Imperceptible information such as velocities and internal forces are discarded from observation. Our goal is to learn an neural-network encoder $f^\theta$ capable of computing a low dimensional latent state $h_t$ from observation: $h_t = f^\theta(o_t)$, where $h_t$ needs to capture sufficient information related to dynamics from the present observation so that it can be used to measure whether two latent states are dynamically similar or not. In particular, two latent states $h_t$ and $h_t'$ are regarded as dynamically equivalent if given an arbitrary action sequence $a_t,a_{t+1}...,a_{t+k}$, their future states $s_{t+k}, s'_{t+k}$ are the same.

\section{Approach}
\label{sec:approach}
The framework of the DiffSRL consists of three main components: 1) a state-observing AutoEncoder which compresses the observation $o_t$ to a latent state $h_t$ and reconstructs the observable state $s_t^\text{obs}$ from the latent state;
2) a constraint regulator which regulates the states against system dynamical constraints as well as compute additional constraint violation loss to regulate the autoencoder;
3) a differentiable simulator to roll out the trajectory starting from the decoded state. 
The rolling-out process is differentiable so that gradients can be propagated from the end of the trajectory all the way back to the encoder. The overall pipeline of DiffSRL is shown in \cref{fig:model}.

In this section, we will first introduce the background related to our design, then move on to details of the three major components and our loss function. \textcolor{black}{Finally, we illustrate more details on implementing of our framework on real robots.}

\subsection{Preliminaries}

\noindent\textbf{Particle-In-Cell paradigm:} Lagrangian methods and Euler methods are two major branches of simulation algorithms~\cite{MPM}. The former carries physical properties with particles and is convenient for numerical integration, while the latter maintains physical properties on fixed grids and can be used for fast and interactive simulation. The Particle-In-Cell (PIC) method combines both the methods and is commonly applied in fluid and deformable-object simulation. The deformable-object simulation in the PlasticineLab is based on the Material Point Method~\cite{MPM}, a variant of Particle-In-Cell. Our constraint regulator utilizes the PIC paradigm to detect and resolve penetration efficiently.


\noindent\textbf{Distance Metric on Point Clouds:} Point clouds usually are expressed as unordered sets. Hence, differences between two point clouds cannot be measured by common metrics such as MSE or Mean Average Error (MAE). The Earth Mover’s Distance (EMD)~\cite{emd} is adopted in the DiffSRL to compare two point clouds in terms of the minimum total pairwise distance. The EMD can be expressed as:
\begin{align}
    \text{Dist}_\text{emd}(A,B) = \min_{\phi:A\rightarrow B} \sum_{a\in A} d(a, \phi(a))
    \vspace{-5pt}
    \label{eq:emd_dist}
\end{align}
where $A$ and $B$ are two unordered point sets, $d(\cdot, \cdot)$ is the distance between two sets, and $\phi(\cdot)$ is the optimal correspondence between $a$ and $b$ which is computed iteratively by the auction algorithm. The EMD has been made differentiable thanks to works such as \cite{emd} in 3D vision.
\subsection{State Observation Autoencoder}
\label{subsec:autoencoder}
To obtain the latent states of a dynamic system, a deep neural network with learnable parameters $\theta_\text{encoder}$ is used as the encoder, where the specific architecture depends on the type of observations. For instance, the MLP is applied on ordered vector observation, Convolution Neural Network (CNN) on images, and permutation invariant encoders such as PointNet~\cite{PointNet} on unordered sets including point clouds. Meanwhile, a decoder with parameters $\theta_\text{decoder}$ is trained simultaneously to recover information from latent states. Formally, the AutoEncoder can be expressed as:
\begin{align}
    h_t = f^{\theta_\text{encoder}}(o_t),\ \  
    \Tilde{s}^\text{obs}_t = f^{\theta_\text{decoder}}(h_t)
\end{align}
where $o_t, h_t$ are as defined in \cref{sec:problem} and $\Tilde{s}^\text{obs}_t$ is the reconstructed observation from $h_t$.

It deserves notice here that since the observation $o_t$ is selected from only the observable part in state $s_t$, the outcome from the decoder, $\Tilde{s}^\text{obs}_t$, is expected to contain no imperceptible data as well. Thus, before this outcome enters the rest components, the imperceptible information needs to be appended back to $\Tilde{s}^\text{obs}_t$ to formulate the $\Tilde{s}_t$. This procedure can be expressed as:
\begin{align}
    \Tilde{s}_t &= s_t \ominus s^\text{obs}_t \oplus \Tilde{s}^\text{obs}_t
    \label{eq:reconstructed_state_formulation}
\end{align}
where $s^\text{obs}_t$ denotes the observable part of a state. Hence, $s_t \ominus s^\text{obs}_t$ gives the imperceptible part of the state, and the $\oplus$ operation adds the reconstructed $\Tilde{s}^\text{obs}_t$ back.

\subsection{Differentiable Constraint Projector}

The reconstructed state $\Tilde{s}_t$ may violate hard dynamics constraints, such as non-interpenetration constraints, joint limit constraints, and continuity constraints within continuous deformable objects. Constraint violation usually causes meaningless gradient,  significant numerical errors or even simulator failure. Hence, it is necessary to make sure the input to the differentiable simulator is always valid as well as provide incentive for the AutoEncoder’s to generate feasible state throughout the training procedure. Therefore, a constraints regulator is designed to project decoder output state to the closest the feasible state and provide incentive for the AutoEncoder to stay away from unrealistic states. Formally, the reconstructed state, the constraint regulator loss, and the projected state are computed as:
\begin{align}
    \mathcal{L}_\text{constraint} &= \min_{s\in \mathcal{S}} \text{Dist}_\text{emd}(s,\Tilde{s}_t), \label{eq:constraint-loss} \\
    \hat{s}_t &= \arg\min_{s\in \mathcal{S}} \text{Dist}_\text{emd}(s,\Tilde{s}_t),
\end{align}
\textcolor{black}{where the distance $\text{Dist}$ is the aforementioned EMD \cref{eq:emd_dist} between point clouds.} We will explain details of how to compute the closest feasible state in \ref{sec:exp} as well as \ref{alg: regulator}

\subsection{Differentiable Simulator and loss design}

By using the reconstructed state from decoder, the simulation in \cref{eq:sim_dynamics} then becomes
\begin{align}
    \hat{s}_{t+1} &= f_\text{sim}(\hat{s}_t,a_t)
\end{align}
where $\hat{s}_t$ is output state of decoder after constraint projection, from which the simulator $f_{\text{sim}}$ executes the input action $a_t$ and reaches the corresponding result state $\hat{s}_{t+1}$. When applied to a trajectory of length $k$, a sequence $\hat{s}_{t:t+k}$ can be thereafter retrieved starting from $\hat{s}_t$, along an action sequence $a_t, a_{t+1}, ..., a_{t+k}$. 

The model is trained with both the constraints violation loss $\mathcal{L}_{\text{constraint}}$ (\cref{eq:constraint-loss}) and the multistep reconstruction loss $\mathcal{L}_{\text{multi-step}}$ defined as
\begin{align}
    \mathcal{L}_{\text{multi-step}} = \sum_{i=1}^k\gamma^i\text{Dist}_\text{emd}(s_{t+i},\hat{s}_{t+i}),
    \label{eq:loss}
\end{align}
\textcolor{black}{The multistep reconstruction loss penalizes the distance between each state $s_{t+i}$ in the reference trajectory from the dataset and its correspondence in the rolling-out trajectory $\hat{s}_{t:t+k}$.} A decaying factor $\gamma$ is used to mitigate the gradient instability effect along backpropagation through a long horizon. In this way, the total loss to be optimized is
\begin{align}
    \mathcal{L}_{\text{total}} = (1-\beta)\mathcal{L}_{\text{multi-step}} + \beta\mathcal{L}_{\text{constraints}}
    \label{eq:multiobj}
\end{align}
where the weight factor $\beta$ is introduced to trade-off among different loss terms during training. Initially, the decoded state violates constraints significantly, so $\beta$ is set to $1$ to encourage the AutoEncoder to respect the physical constrains. Progressively, $\beta$ decays exponentially at rate $\lambda$ so that the multistep reconstruction loss dominates.

\subsection{Implementation Details}

\begin{figure}[b!]
\centering
  \vspace{-20pt}
  \includegraphics[width=0.8\linewidth]{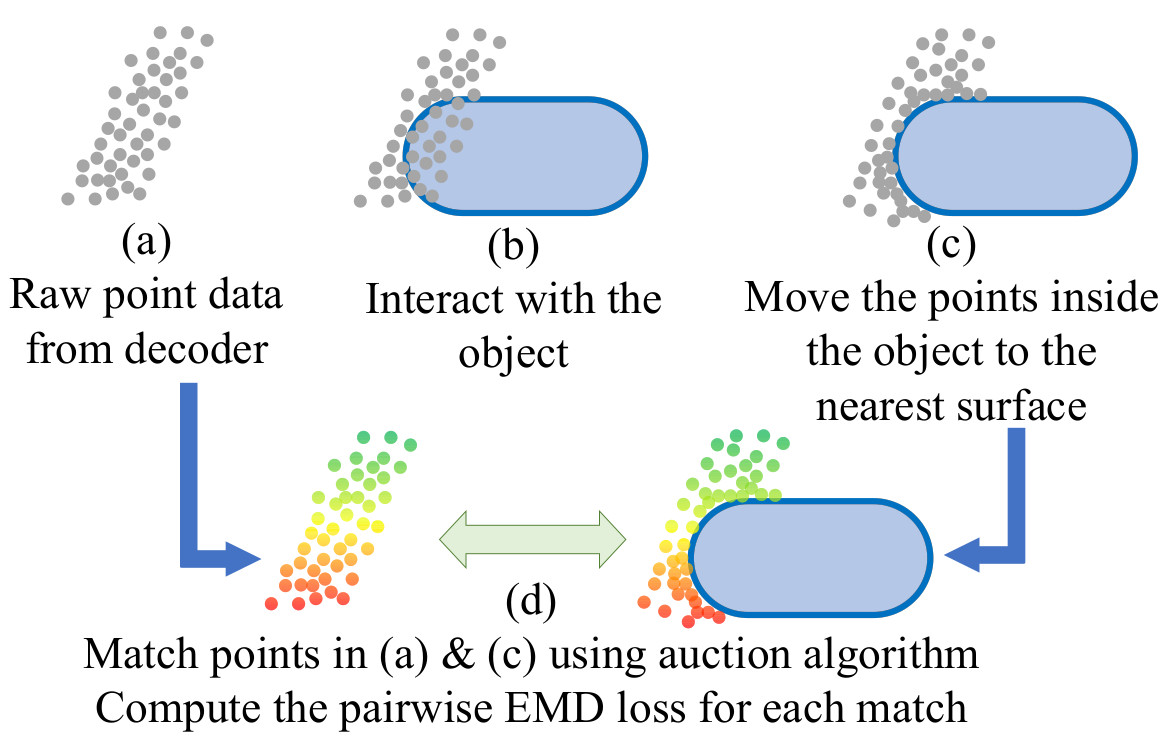}
  \caption{The collision-free and smooth-velocity-field constraint projector, where the dots and the capsule represent the point clouds and the rigid-body object which may collide with them, respectively.}
  \label{fig:Regulator}
\vspace{-5pt}
\end{figure}

The MPM of the PlasticineLab provides a direct access to the particles that composes the deformable objects. Therefore, as is declared in \cref{sec:problem}, we set each observation $o_t$ as a subset of the positions of all particles, which can be naturally considered as an unordered point cloud. \textcolor{black}{Hence, a variant of Point Completion Network (PCN)~\cite{PCN} which is an extension of the PointNet ideology, is used to implement the AutoEncoder.} 

As for the constraint projector, the MPM simulation disallows penetration between particles and rigid bodies and also requires smooth velocity fields within continuous materials. Hence, our projector is constructed in two stages:
1) We use signed distance field (SDF) with MPM to efficiently detect penetration. For those particles $p_\text{collide}$ who have negative signed distance and reside inside rigid bodies, we resolve the penetration by finding the minimum distance $d_{p_\text{collide}}$ to exit the colliding object. For simplicity, it is assumed that each rigid-body object is a geometric primitive and has an analytical collision detection routine so that the minimum exit distance can be solved efficiently using a simple linear programming solver. The loss function for this objective is defined as the sum of all $d_{p_\text{collide}}$.
2) We use auction algorithm\cite{auction}, a GPU parallelizable point correspondence matching algorithm, to find the best pairwise matches between particles in reference state and those in the reconstructed state after collision resolution in 1). The step is necessary because, as is emphasized in \cref{subsec:autoencoder}, the imperceptible information such as velocity and shearing force of each particle is still missing in the decoder's outputs $\Tilde{s}^\text{obs}_t$. The information needs to be consistent with field properties maintained by the grid to execute the MPM properly. After the auction algorithm establishes pairwise relationships, each ground-true particle's properties are exerted to its match, which fulfills the $\oplus$ notation in \ref{eq:reconstructed_state_formulation}. The details are described in \cref{alg: regulator} and also illustrated in \cref{fig:Regulator}. The pairwise distance between the reference state and projected state is also added as part of constraint satisfaction loss.

\begin{algorithm}
\caption{Constraints Projector}\label{alg: regulator}
\begin{algorithmic}
\State \textbf{Input:} Rigid bodies $\mathcal{R}$, the ground-truth particle observable state $s^\text{obs}_t$ (i.e., position of particles), and decoded particles position $\Tilde{s}^\text{obs}_t$
\State \textbf{Initialize:} Grids to store interpenetrating rigid bodies information $c_\text{grid}^\text{col}$, loss of penetration $l_\text{penetration}$, and loss reconstruction $l_\text{rec}$.
\State $\triangleleft$ $\hat{s}_t^\text{obs, no-penetration} \gets \text{ResolvePenetration}(\mathcal{R},\Tilde{s}^\text{obs}_t)$
\State $\triangleleft$ Compute grid mass using particle-in-cell
\State $m_\text{grid} \gets \text{ComputeGridMass}(\Tilde{s}^\text{obs}_t)$
\For{$r$ in $\mathcal{R}$}
    \State $\triangleleft$ Check penetration using signed distance field
    \State $c_\text{grid}^\text{col} \gets \text{SDF}(r,m_\text{grid})$
\EndFor
\State $\triangleleft$ Compute Interpenetration related information of each particle to all rigid bodies
\For{$p$ in $\hat{s}_t^{\text{obs no-penetration}}$}
    \If{p.grid is not empty}
        \State $\triangleleft$ Solve linear programming problem to find minimum displacement for a particle to stay out of penetrating rigid body
        \State $\text{p.pos} \gets \text{LPSolver}(\mathcal{R},p)$ 
        \State $\triangleleft$ Sum up minimum displacement costs as loss
        \State $l_\text{penetration} \mathrel{+}= \text{LPCost}(r,p)$
    \EndIf
\EndFor
\State $\triangleleft$ $\hat{s}^\text{obs}_t \gets \text{ResolveMissMatch}(\hat{s}_t^\text{obs no-penetration})$
\State $\text{order} \gets \text{AuctionAlgorithm}(s^\text{obs}_t,\hat{s}^\text{obs}_t)$
\State $\hat{s}^\text{obs}_t \gets \text{perm}(\hat{s}_t^\text{obs no-penetration}, \text{perm(order)})$
\State return $\hat{s}^\text{obs}_t$
\end{algorithmic}
\end{algorithm}

\subsection{Closing the Sim2Real Gap}

In the simulator, the observable part of states $s_t^\text{obs}$ are in the form of dense point clouds. However, a real-world sensor, such as a depth camera and a LiDAR, can only provide a thin-shell of noisy point cloud of an object's surface. Uniformly sampling new points within the interior of thin-shell can produce a dense point clouds filling the object's volume, but such point clouds may observe a different distribution compared with those in the simulator. \textcolor{black}{Thus, besides uniformly sampling new points within the thin-shell when the algorithm is tested on real robot, we augment the dataset by adding a random normal noise to point positions and randomly removing 10\% - 50\% particles from the dense point cloud during the training in the simulator.}

\section{Experiments and Results}
\label{sec:exp}

To assess the effectiveness of the DiffSRL, we evaluated the model with four experiments:
1) trajectory reconstruction,
2) reward prediction,
3) KNN-EMD, and
3) model-free reinforcement learning on 6 single object manipulation tasks from the PlasticineLab benchmark~\cite{PlasticineLab}. The proposed DiffSRL has achieved state-of-the-art performance in these experiments most of the time.

\subsection{Baselines for Comparison}
We compared our approach against state-of-the-art method CFM~\cite{CFM} and four commonly used dynamic state representation learning methods: the \textit{E2C} method~\cite{E2C}, the \textit{Forward} method~\cite{Forward} that learns a dynamic forward model on latent states, the \textit{Inverse} method~\cite{Inverse} that conducts inverse-model learning for action prediction, as well as the \textit{AutoEncoder} method directly reconstructing the state ~\cite{AutoEncoder}. In \cref{tab:baselines}, we summarize characteristics of different approaches. 

Regarding implementation details of these baselines, similar to the DiffSRL, we used EMD~\cite{emd} as a loss function to train the AutoEncoder in the baseline. The forward model used in the E2C and the Forward $f^\theta_\text{fwd}(h_t, a_t)$ are 3-layer MLP on latent space. The action predictor $\hat{a}_t = f^\phi_\text{inverse}(h_t,h_{t+1})$ in the Inverse uses a pair of successive latent states as inputs and tried to infer the action with a 3-layer MLP. The contrastive loss in CFM~\cite{CFM} uses the same formation and negative sample selection method as in the original paper. For all MLPs mentioned here, we used 256 as the hidden-layer size.

\begin{table}[t]
\centering
\small
\resizebox{\linewidth}{!}{ 
\begin{tabular}{ p{2.1cm} | m{1.1cm} | m{0.8 cm} | m{1.4cm} | m{1.2cm} | m{1.2cm} }
\toprule
                                & predictive dynamic model  & predict action   & observation reconstruction & multistep information  & physical constraints\\ \midrule
\hline
DiffSRL (Our)                   &  $\surd$             &                   & $\surd$                        &   $\surd$                               & $\surd$                      \\
\hline
E2C\cite{E2C}                   &  $\surd$             &                   & $\surd$                        &                                         &                              \\
\hline
Forward\cite{Forward}               &  $\surd$             &                  &                                &                                         &                              \\
\hline
Inverse\cite{Inverse}            &                     &  $\surd$          &                                &                                         &
 \\
\hline
CFM\cite{CFM}                   &  $\surd$             &                   &                                &                                         &                              \\ 
\hline
Autoencoder\cite{AutoEncoder}   &                      &                   & $\surd$                        &                                         &                              \\
\bottomrule
\end{tabular}
}
\caption{Comparison of different components used in different state representation learning pipelines.}
\label{tab:baselines}
\vspace{-20pt}
\end{table}

\subsection{ Environment selection and Data Collection}
We selected 6 environments from PlasticineLab, each involving one single deformable object. As for \textit{TripleMove}, \textit{table}, \textit{Assemble}, these experiment involves multiple deformable objects and require special design of loss functions as well as adaptation. We left handling multi-objects as future work.
\textcolor{black}{For each environment, we collect 6,000 trajectories of 8-step length from simulator.}
30\% of data were sampled from random policy, while others were collected when the model optimized trajectories for manipulation. The reason for doing so is that random actions frequently cause manipulators detaching from the deformable objects too early to induce sufficient deformation. We use the ratio of 8:1:1 to split collected samples into the training, validation and testing sets.



\begin{table*}[t!]
\centering
\resizebox{0.99\linewidth}{!}{
\begin{tabular}{@{}lrrrrrrrrrrrrrrrrrr@{}}
\toprule
\textbf{Datasets} &
  \multicolumn{3}{c}{Rope} &
  \multicolumn{3}{c}{Chopsticks} &
  \multicolumn{3}{c}{Torus} &
  \multicolumn{3}{c}{Writer} &
  \multicolumn{3}{c}{Rollingpin} &
  \multicolumn{3}{c}{Pinch}\\ \midrule
\textbf{Metric} &
  \multicolumn{1}{c}{EMD} &
  \multicolumn{1}{c}{MSE} &
  \multicolumn{1}{c}{KNN} &
  \multicolumn{1}{c}{EMD} &
  \multicolumn{1}{c}{MSE} &
  \multicolumn{1}{c}{KNN} &
  \multicolumn{1}{c}{EMD} &
  \multicolumn{1}{c}{MSE} &
  \multicolumn{1}{c}{KNN} &
  \multicolumn{1}{c}{EMD} &
  \multicolumn{1}{c}{MSE} &
  \multicolumn{1}{c}{KNN} &
  \multicolumn{1}{c}{EMD} &
  \multicolumn{1}{c}{MSE} &
  \multicolumn{1}{c}{KNN} &
  \multicolumn{1}{c}{EMD} &
  \multicolumn{1}{c}{MSE} &
  \multicolumn{1}{c}{KNN} \\ \midrule
AutoEncoder   & 0.048 & 0.236 & 0.830 & 0.044 & 1.102 & \underline{0.350} & 0.049 & \underline{0.011} & \underline{0.021} & 0.057 & 0.005 & 0.169 & 0.037 & 0.019& \textbf{0.103} & 0.031 & 0.005 & \underline{0.071}\\
Inverse       & 0.042 & 0.276 & 1.043 & 0.048 & 1.190 & 0.449 & 0.041 & 0.099 & 0.028 & 0.067 & 0.008 & 0.017 & 0.032 & 0.093 & 0.158 & 0.026 & 0.046 & 0.131\\
Forward       & 0.083 & 5.435 & \underline{0.816} & 0.124 & 6.599 & 0.357 & 0.192 & 0.085 & 0.021 & 0.214 & 0.007 & 0.016 & 0.091 & 0.738 & 0.112 & 0.512 & 0.039 & 0.075\\
E2C           & 0.048 & \underline{0.223} & 0.915 & 0.064 & 1.244 & 0.359 & 0.039 & 0.098 & 0.026 & 0.058 & 0.008 & \textbf{0.015} & 0.034 & 0.032 & 0.153 & 0.027 & 0.043 & 0.166\\
CFM           & \underline{0.037} & 0.236 & 1.137 & 0.037 & \underline{1.010} & 0.491 & \textbf{0.029} & 0.012 & 0.032 & 0.078 & \underline{0.004} & 0.036 & 0.038 & \underline{0.015} & 0.124 & 0.026 & \underline{0.004} & 0.161 \\\midrule
DiffSRL       & \textbf{0.027} & \textbf{0.218} & \textbf{0.797} & \textbf{0.031} & \textbf{0.840} & \textbf{0.324} & \underline{0.036} & \textbf{0.009} & \textbf{0.020} & \textbf{0.044} & \textbf{0.003} & \underline{0.016} & \textbf{0.021} & \textbf{0.013} & \underline{0.103} & \textbf{0.016} & \textbf{0.003} & \textbf{0.071}\\
$\text{DiffSRL}^-$ & 0.039 & 0.292 & 0.821 & \underline{0.036} & 1.226 & 0.376 & 0.038 & 0.012 & 0.023 & \underline{0.051} & 0.006 & 0.018 & \underline{0.028} & 0.017 & 0.121 & \underline{0.021} & 0.004 & 0.091\\\bottomrule
\end{tabular}}
\caption{Comparison of the average of trajectory roll-out earth mover's distance and mean squared error of reward prediction as well as KNN-MEMD. The smaller the MSE and EMD, the better the model.
The best result is highlighted in \textbf{bold} and the second best is highlighted with \underline{underline}. $\text{DiffSRL}^-$ is our proposed method trained without non-penetration constraint projector.}
\label{tab:results}
\vspace{-20pt}
\end{table*}

\subsection{Comparison in Trajectory Reconstruction and Reward Prediction}

The trajectory reconstruction error and the reward prediction error are two straightforward metrics to reflect the encoder's capability of dynamics awareness. The trajectory reconstruction error measures the similarity between the reference trajectory and the trajectory rolled-out from the reconstructed state given the same action sequence, which is computed by accumulating the earth mover's distance~\cite{emd}, i.e., $\sum_{s_{t:t+k}}D_\text{EMD}(s_t,\hat{s}_t)$, between each pair of states in both trajectories. Some baseline methods such as \textit{CFM}, \textit{Forward} and \textit{Inverse} only train the encoder without a decoder and thus cannot reconstruct states. Thus, we trained a decoder for each fixed encoder using the AutoEncoder training pipeline until convergence. \cref{tab:results} shows that DiffSRL achieved the best performance among all models. 

The reward prediction error evaluates whether an encoded latent state can capture enough information from a state to accurately predict reward for a given task. For each task, we trained a 3-layer MLP for latent states obtained from each encoder to predict rewards from latent states. The outputs of the MLP are the rewards related to the poses of deformable objects. The error is measured using mean square error: $\frac{1}{|\mathcal{D}|}\sum_{s\in |\mathcal{D}|}\|\hat{r}_\text{pred}-r_\text{real}\|^2$. As summarized in \cref{tab:results}, our model achieved the best performance on most of datasets. One thing worth noting is that the \textit{Forward} model observes the worst performance on both metrics among all models, probably because it tends to be trapped in a trivial minimum as mentioned in \cite{SRLSurvey}.

\begin{figure}[!h]
\centering
\vspace{-5pt}
  \includegraphics[width=0.7\linewidth]{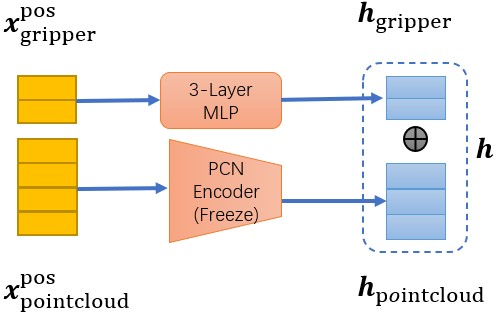}
  \vspace{-5pt}
  \caption{The architecture of policy network that we used in model-based policy optimization and model-free reinforcement learning.}
  \label{fig:policynet}
  \vspace{-20pt}
\end{figure}

\subsection{KNN-MEMD}
KNN-MSE\cite{KNN-MSE} has been used as a metric for static state representation learning. It can evaluate whether neighbors on latent space are actually close to each other on state space. For dynamic state representation, we need not only check whether two states from a neighborhood are similar in current time stamp, but also future states driven by the same action sequence. Moreover, since two states are two point clouds in our case, we replaced the MSE with the Mean Earth Mover's Distance(MEMD) as the distance metric. Predictably, dynamical equivalent states will have a smaller KNN-MEMD. Our novel metrics is evaluated as follow:
\begin{align}
\text{KNN-MEMD}(h) = \frac{1}{K}\sum_{h'\in \text{knn}(h,K)}\frac{1}{H}\sum_t^H \text{EMD}(s_t,s'_t)
\end{align}
H is the rollout horizon, K is number of neighbors, $h,h'$ are latent states and $s,s'$ are states. \ref{tab:results}

\subsection{Model-Free Reinforcement Learning (MFRL)}

\begin{figure*}[t!]
\centering
  \includegraphics[width=0.8\linewidth]{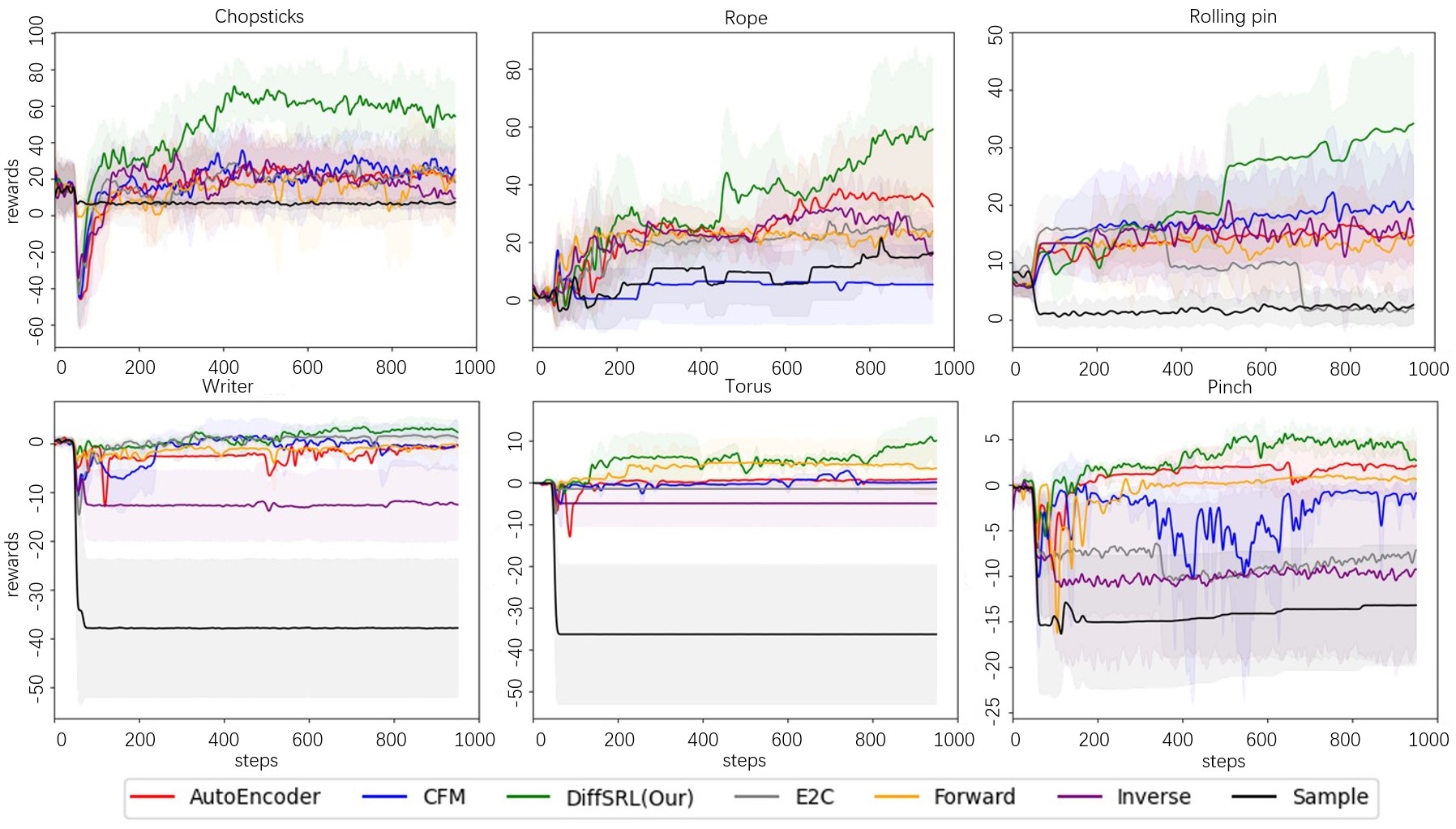}
  \vspace{-5pt}
  \caption{TD3 Reward in different environments.}
  \label{fig:rewards}
\vspace{-10pt}
\end{figure*}

\begin{figure*}[ht!]
\centering
  \includegraphics[width=0.8\linewidth]{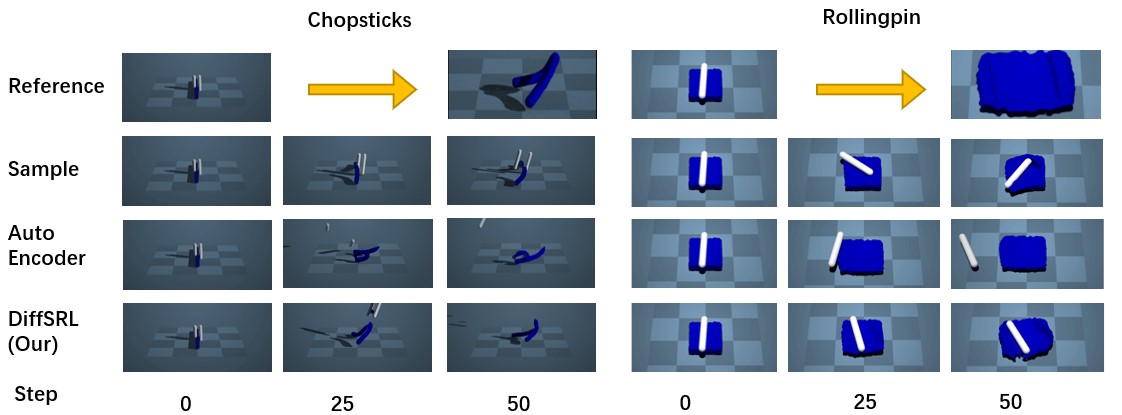}
  \caption{Keyframes of different approaches in MFRL task.
  }
  \vspace{-15pt}
  \label{fig:keyframe}
\end{figure*}

To study the effectiveness of state representation learning on various complex downstream tasks, we evaluated encoders trained from different methods on training model free reinforcement learning. We used TD3, a state-of-the-art off policy method ~\cite{TD3} to train policies based on latent state observation. The trained policy is expected to control manipulators to move the plasticine to different target positions within finite steps. Detailed target and reward design were the same as described in \cite{PlasticineLab}. We used latent states for both the state-value function and the policy function. The architecture of the policy network is shown in \cref{fig:policynet}, where the encoder's weights are fixed during policy learning. For comparison, we added another baseline: the naive down-sampling based method provided by PlasticineLab, which sampled 400 points from each point cloud\footnote{The original down-sampling based method from the PlasticineLab is not feasible in practice since it requires accurately tracking 400 points within the deformable object,which is impractical.}. The policy network is initialized as a simple MLP, and its inputs are obtained by concatenating features of these 400 particles. We evaluated performance using the per-epoch average rewards. Each experiment was repeated for five times, and the means and standard deviations are plotted in \cref{fig:rewards}. From the first row we can see that most state representation learning methods can help policy improving faster and achieve better final rewards compared with the down-sampling based method. Moreover, our approach achieved the best result among all state representation learning methods, including the \textit{AutoEncoder} that was trained using a similar loss function. Some keyframes from the trajectory are shown in \cref{fig:keyframe}.

\begin{figure}[t!]
\centering
  \includegraphics[trim=0 0 0 40, clip, width=0.70\linewidth]{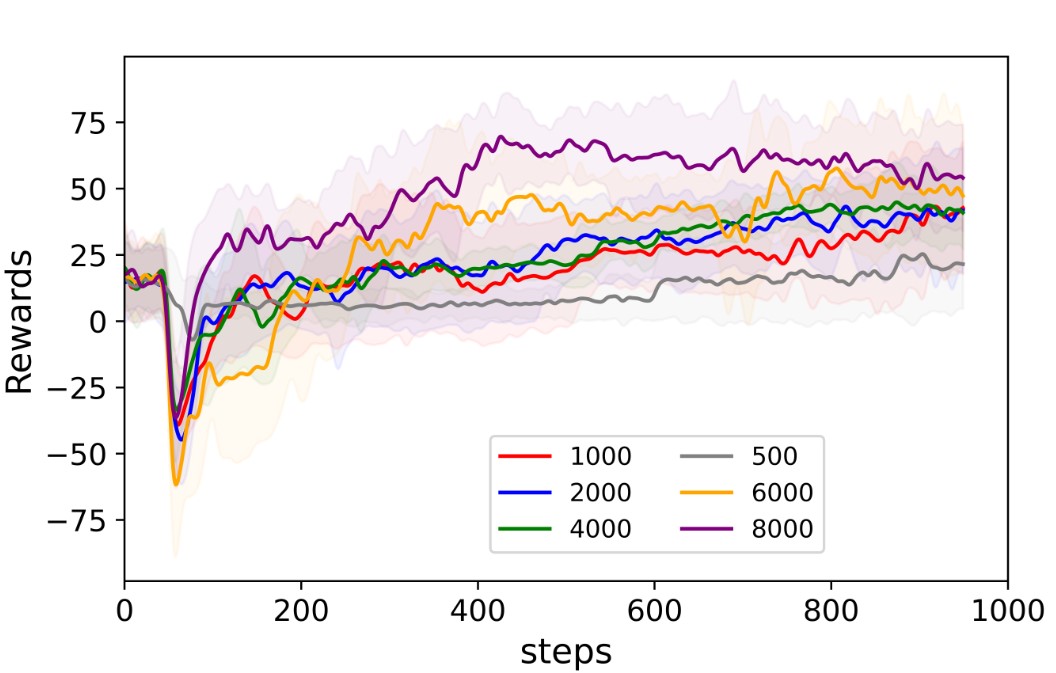}
  \caption{MFRL reward in Chopsticks when using different number of point as encoder input. Averaging over 5 experiments for each input particle number.}
  \label{fig:robustness}
\vspace{-12pt}
\end{figure}


\subsection{Ablation Study}

\begin{figure}[h!]
\centering
  \includegraphics[width=\linewidth]{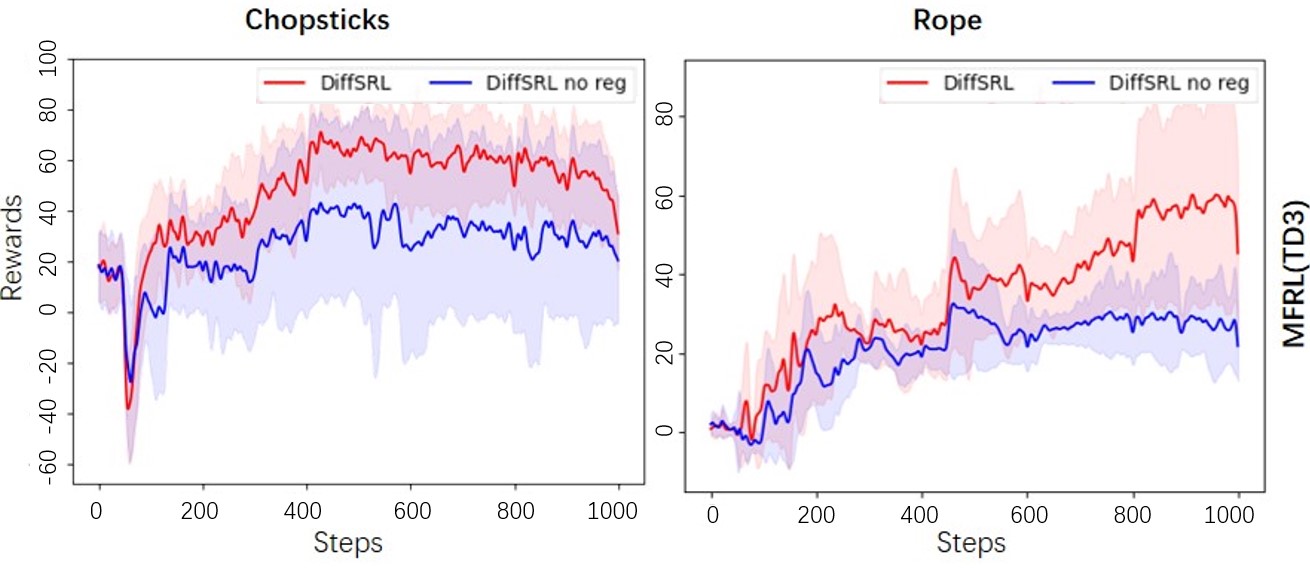}
  \caption{Ablation Study: without non-penetration projector}
  \label{fig:ablation}
  \vspace{-20pt}
\end{figure}

\textcolor{black}{To evaluate the effectiveness of the constraint projector, we conducted an ablation study by removing non-penetration projector during the training. We compared its results with those of our method both quantitatively on different metrics as shown in \ref{tab:results} and qualitatively using MFRL training as in \ref{fig:ablation}. Regarding the trajectory reconstruction errors, the reward prediction errors and the KNN-MEMD, it shows that removing the non-penetration projector will result in performance degradation. We also tried to remove the smoothness projector during training, but this will lead to numerical issues in the differentiable simulator because it violates MPM's requirement that particles in the same grid must have consistent properties.
Furthermore, the encoders without non-penetration constraints projector shows worse results in reinforcement learning. Therefore, we can conclude that constraints projector is essential for improving performance.}


\subsection{Robustness Analysis}
The number of points observed by a sensor may vary significantly in real applications. Since the PCN-based point cloud encoder can be directly applied with different point number without re-training, we used various numbers of observed points to test the robustness of our model. We trained a DiffSRL model using point clouds consisting of 8,000 points and then investigated the deployment performance using 6,000, 4,000, 2,000 particles by computing the reward curves when training MFRL in latent space. As shown in \cref{fig:robustness}, although using fewer observable particles (from 8,000 to 6,000) can decrease the performance, further decreasing of the particle number (from 6,000 to 1,000) will not significantly downgrade the performance, which demonstrates that our method is relatively robust against disturbance in observation. Only we set number of observable particles to 500 , can we observe a significant decay in down stream task's performance.

\subsection{Hyperparameters Setting}
\begin{figure}[tb!]
\centering
  \includegraphics[trim=0 0 0 0, clip, width=0.75\linewidth]{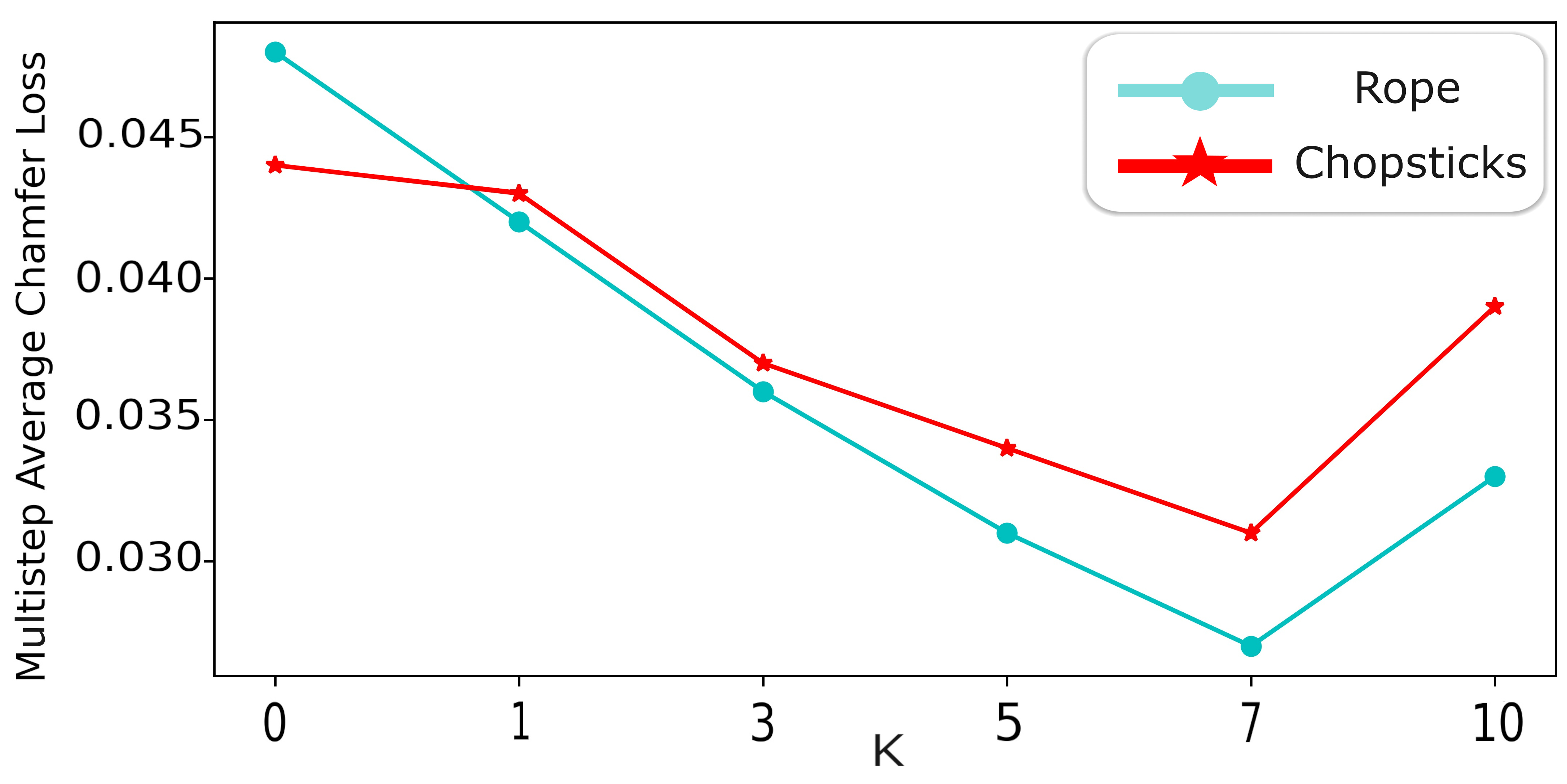}
  \vspace{-5pt}
  \caption{10-step EMD loss on both rope and chopstick when using different of training steps.}
  \label{fig:training_step}
  \vspace{-5pt}
\end{figure}

\begin{table}[t!]
\small
\centering
\resizebox{0.8\linewidth}{!}{ 
\begin{tabular}{lrr}
\toprule
Parameters          &  Meaning              & Value  \\ \midrule
N                   &  Number of particles  & 8192   \\
I                   &    ICP iterations     & 3000   \\
$d_\text{latent}$   & Latent Space dimension& 1024   \\
$\alpha$            & Learning rate         & 1e-5   \\
$\gamma$            & weight decay rate in \ref{eq:loss}  & 0.99   \\
$\beta$             & Initial weight between in \ref{eq:multiobj}  & 0.99   \\
$\lambda$           & Decay rate of $\beta$ per epoch         & 0.9    \\
E                   & Number of epochs             & 20     \\
 \bottomrule
\end{tabular}
}
\caption{Hyperparameters}
\label{tab:params}
\vspace{-5pt}
\end{table}

\cref{tab:params} summarizes the hyperparameters used in our model. One important hyperparameter of our method is $k$, the number of trajectory rollout steps. We chose $k$ by evaluating the 10-step trajectory reconstruction loss on the validation set, with the result shown in \cref{fig:training_step}. For both environments, $k = 7$ achieved the best performance. Notice that the loss curve demonstrates non-monotonic behavior as the training step size increases, and one possible reason is that with a larger step size, the gradient back-propagation through time tends to be noisier due to numerical issues, and the optimization landscape becomes more wiggly. We leave a further investigation as future work.

\subsection{Hardware Experiment}
\begin{figure}[tb!]
\centering
  \includegraphics[width=\linewidth]{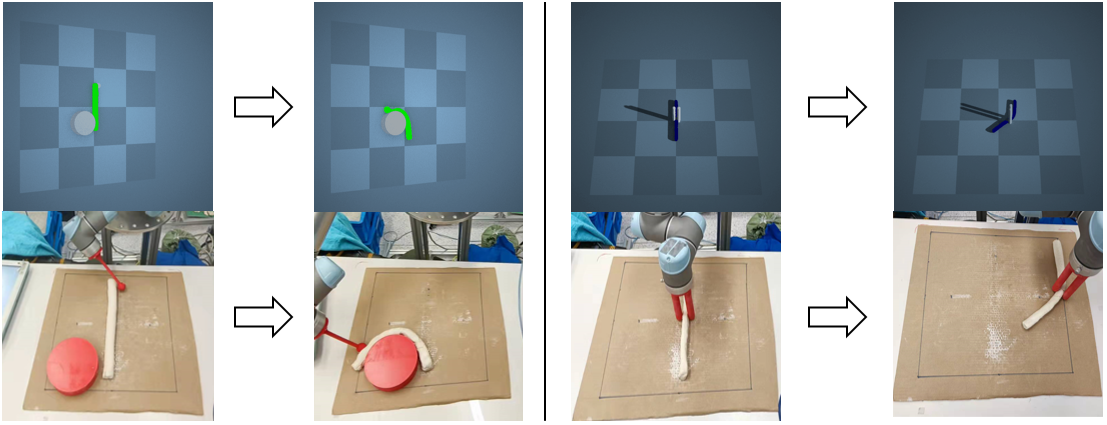}
  \caption{Simulated and real-robot experiments for Rope and Chopsticks.}
  \label{fig:realrobot}
  \vspace{-20pt}
\end{figure}

We deployed both policies trained with MFRL on a UR5 robot with a Kinect-v2 depth camera. We set up two tasks that required the robotic arm to manipulate a plasticine on a plane as shown in \cref{fig:realrobot}. We first segment the plasticine from the background using color. Next, to obtain a dense point cloud for the volume taken by the plasticine, we use the points on the top of the plasticine's surface as seeds. For each seed we sampled more points below it to infill the volume. \cref{fig:realrobot} also shows the start and terminal states of these two tasks both in simulation and real world. As demonstrated, our method can work well on real robots. We found that the occlusion could be an issue and thus we designed the end-effector to be thin and tall to minimize the possible occlusion. How to handle the partial observation will be left as future work.

\section{Conclusion and Future Work}
We present a novel dynamic state representation learning model for deformable objects. Our model has been evaluated with multiple tasks on soft-body manipulation in PlasticineLab~\cite{PlasticineLab} and its performance has been compared with other state-of-the-art models.
Meanwhile, the applicability of the method on real-world scenarios has been examined using the robotic arms. One possible follow-up might include the differentiable rendering method~\cite{diffrender} in our pipeline, which enables using more accessible images as observation while maintaining the overall differentiability.
\label{sec:conclusion}

{\small
\bibliographystyle{IEEEtran}
\bibliography{references}

\begin{thebibliography}{10}
\providecommand{\url}[1]{#1}
\csname url@rmstyle\endcsname
\providecommand{\newblock}{\relax}
\providecommand{\bibinfo}[2]{#2}
\providecommand\BIBentrySTDinterwordspacing{\spaceskip=0pt\relax}
\providecommand\BIBentryALTinterwordstretchfactor{4}
\providecommand\BIBentryALTinterwordspacing{\spaceskip=\fontdimen2\font plus
\BIBentryALTinterwordstretchfactor\fontdimen3\font minus
  \fontdimen4\font\relax}
\providecommand\BIBforeignlanguage[2]{{%
\expandafter\ifx\csname l@#1\endcsname\relax
\typeout{** WARNING: IEEEtran.bst: No hyphenation pattern has been}%
\typeout{** loaded for the language `#1'. Using the pattern for}%
\typeout{** the default language instead.}%
\else
\language=\csname l@#1\endcsname
\fi
#2}}

\bibitem{E2C}
M.~Watter, J.~T. Springenberg, J.~Boedecker, and M.~A. Riedmiller, ``Embed to
  control: {A} locally linear latent dynamics model for control from raw
  images,'' in \emph{NeurIPS}, 2015.

\bibitem{AutoEncoder}
C.~Finn, X.~Y. Tan, Y.~Duan, T.~Darrell, S.~Levine, and P.~Abbeel, ``Deep
  spatial autoencoders for visuomotor learning,'' in \emph{ICRA}, 2016.

\bibitem{CFM}
W.~Yan, A.~Vangipuram, P.~Abbeel, and L.~Pinto, ``Learning predictive
  representations for deformable objects using contrastive estimation,'' in
  \emph{{CoRL}}, 2020.

\bibitem{G-DOOM}
X.~Ma, D.~Hsu, and W.~S. Lee, ``Learning latent graph dynamics for deformable
  object manipulation,'' in \emph{{CoRL}}, 2021.

\bibitem{SRLSurvey}
T.~Lesort, N.~D. Rodr{\'{\i}}guez, J.~Goudou, and D.~Filliat, ``State
  representation learning for control: An overview,'' \emph{Neural Networks},
  2018.

\bibitem{Inverse}
``Learning to poke by poking: Experiential learning of intuitive physics,'' in
  \emph{NeurIPS}, 2016.

\bibitem{Duan2017}
W.~Duan and K.~Jens, ``Learning state representations for robotic control:
  Information disentangling and multi-modal learning,'' Master's thesis, TU
  Delft, 2017.

\bibitem{LossIsRewards}
E.~Shelhamer, P.~Mahmoudieh, M.~Argus, and T.~Darrell, ``Loss is its own
  reward: Self-supervision for reinforcement learning,'' in \emph{{ICLR}},
  2017.

\bibitem{ICM}
D.~Pathak, P.~Agrawal, A.~A. Efros, and T.~Darrell, ``Curiosity-driven
  exploration by self-supervised prediction,'' in \emph{{CVPR} Workshops},
  2017.

\bibitem{PointNet}
C.~R. Qi, H.~Su, K.~Mo, and L.~J. Guibas, ``Pointnet: Deep learning on point
  sets for 3d classification and segmentation,'' in \emph{CVPR}, 2017.

\bibitem{MPM}
C.~Jiang, C.~A. Schroeder, J.~Teran, A.~Stomakhin, and A.~Selle, ``The material
  point method for simulating continuum materials,'' in \emph{{SIGGRAPH}},
  2016.

\bibitem{nimblephysics}
K.~Werling, D.~Omens, J.~Lee, I.~Exarchos, and C.~K. Liu, ``Fast and
  feature-complete differentiable physics engine for articulated rigid bodies
  with contact constraints,'' in \emph{RSS}, 2021.

\bibitem{DiffHand}
J.~Xu, T.~Chen, L.~Zlokapa, M.~Foshey, W.~Matusik, S.~Sueda, and P.~Agrawal,
  ``An end-to-end differentiable framework for contact-aware robot design,'' in
  \emph{RSS}, 2021.

\bibitem{DiffTaichi}
Y.~Hu, L.~Anderson, T.~Li, Q.~Sun, N.~Carr, J.~Ragan{-}Kelley, and F.~Durand,
  ``Difftaichi: Differentiable programming for physical simulation,'' in
  \emph{ICLR}, 2020.

\bibitem{ChainQueen}
Y.~Hu, J.~Liu, A.~Spielberg, J.~B. Tenenbaum, W.~T. Freeman, J.~Wu, D.~Rus, and
  W.~Matusik, ``Chainqueen: {A} real-time differentiable physical simulator for
  soft robotics,'' in \emph{ICRA}, 2019.

\bibitem{PlasticineLab}
Z.~Huang, Y.~Hu, T.~Du, S.~Zhou, H.~Su, J.~B. Tenenbaum, and C.~Gan,
  ``Plasticinelab: {A} soft-body manipulation benchmark with differentiable
  physics,'' in \emph{ICLR}, 2021.

\bibitem{Taichi}
Y.~Hu, T.~Li, L.~Anderson, J.~Ragan{-}Kelley, and F.~Durand, ``Taichi: a
  language for high-performance computation on spatially sparse data
  structures,'' \emph{{ACM} Trans. Graph.}, 2019.

\bibitem{emd}
Y.~Rubner, C.~Tomasi, and L.~J. Guibas, ``A metric for distributions with
  applications to image databases,'' in \emph{ICCV}, 1998.

\bibitem{PCN}
W.~Yuan, T.~Khot, D.~Held, C.~Mertz, and M.~Hebert, ``{PCN:} point completion
  network,'' in \emph{3DV}, 2018.

\bibitem{auction}
D.~P. Bertsekas, ``A distributed algorithm for the assignment problem,'' MIT,
  Tech. Rep., 1979.

\bibitem{Forward}
L.~Kaiser, M.~Babaeizadeh, P.~Milos, B.~Osinski, R.~H. Campbell, K.~Czechowski,
  D.~Erhan, C.~Finn, P.~Kozakowski, S.~Levine, A.~Mohiuddin, R.~Sepassi,
  G.~Tucker, and H.~Michalewski, ``Model based reinforcement learning for
  atari,'' in \emph{{ICLR}}, 2020.

\bibitem{KNN-MSE}
P.~Sermanet, C.~Lynch, J.~Hsu, and S.~Levine, ``Time-contrastive networks:
  Self-supervised learning from multi-view observation,'' in \emph{{CVPR}
  Workshops}, 2017.

\bibitem{TD3}
S.~Fujimoto, H.~van Hoof, and D.~Meger, ``Addressing function approximation
  error in actor-critic methods,'' in \emph{ICML}, 2018.

\bibitem{diffrender}
H.~Kato, D.~Beker, M.~Morariu, T.~Ando, T.~Matsuoka, W.~Kehl, and A.~Gaidon,
  ``Differentiable rendering: {A} survey,'' \emph{arxiv 2006.12057}, 2020.

\end{thebibliography}
}

\end{document}